\documentclass[final]{arxiv}

\usepackage{times}
\usepackage{epsfig}
\usepackage{graphicx}
\usepackage{amsmath}
\usepackage{amssymb}
\DeclareMathOperator*{\argminA}{arg\,min} 

\usepackage[pagebackref=true,breaklinks=true,colorlinks,bookmarks=false]{hyperref}

\def\arxivPaperID{10420} 

\def\arxivPaperID{10420} 
\def\confYear{arxiv 2021}

\begin{document}

\title{Privacy-Preserving Image Acquisition Using Trainable Optical Kernel}  

\author{Yamin Sepehri$^{\dagger \ddagger}$, Pedram Pad$^\dagger$, Pascal Frossard$^\ddagger$, and L. Andrea Dunbar$^\dagger$\\
$^\dagger$Centre Suisse d’Electronique et de Microtechnique (CSEM) \\
$^\ddagger$École polytechnique fédérale de Lausanne (EPFL) 

}

\maketitle

\begin{abstract}
   Preserving privacy is a growing concern in our society where sensors and in particular cameras are ubiquitous. In this work, for the first time, we propose a trainable image acquisition method that removes the sensitive identity revealing information in the optical domain before it reaches the image sensor. The method benefits from a trainable optical convolution kernel which transmits the desired information while filters out the sensitive information. As the sensitive information is suppressed before it reaches the image sensor, it does not enter the digital domain therefore is unretrievable by any sort of privacy attack. This is in contrast with the current digital privacy-preserving methods that are all vulnerable to direct access attack. Also, in contrast with the previous optical privacy-preserving methods that cannot be trained, our method is data-driven and optimized for the specific application at hand. Moreover, there is no additional computation, memory, or power burden on the acquisition system since this processing happens passively in the optical domain and can even be used together and on top of the fully digital privacy-preserving systems. The proposed approach is generic and adaptable to different digital neural networks, and desired and sensitive content pairs.  We demonstrate our new method for several scenarios such as smile detection as the desired attribute while the gender is filtered out as the sensitive content. We trained the optical kernel in conjunction with two adversarial neural networks where the analysis network tries to detect the desired attribute and the adversarial network tries to detect the sensitive content. We show that this method is able to reduce $65.1\%$ of sensitive content when it is selected to be the gender and as the kernel is optimized, it only causes $7.3\%$ reduction of the desired information content. Moreover, we reconstruct the original faces using the deep image reconstruction method that confirms the ineffectiveness of reconstruction attacks to obtain the sensitive content.
\end{abstract}


\vspace{1pt}
{\section{Introduction}\label{section:introduction}}

Deep Neural Networks have shown a state-of-the-art accuracy for a variety of different vision tasks ~\cite{girshick_rich_2014,simonyan_very_2015,zhang_facial_2014}. Usually in such tasks, the user needs to load a standard image in a digital device which runs the neural network. Preserving privacy in such systems is difficult as the images are rich in sensitive information and loading private images in digital domain may expose user's sensitive data to attacks. Whilst, in most cases, this sensitive information may not be needed to perform the required vision task. For example, to perform a facial attribute detection task such as smile recognition or eye-gaze estimation, a high amount of private information like the identity or gender are loaded on the digital device which is irrelevant to the task. The private information leakage is at greater risk if the user runs the desired deep learning-based task on a cloud server. As state-of-the-art neural networks in vision tasks are generally demanding in terms of computational power, some users share the images from their end devices to the cloud servers which increases the privacy concern due to a possibility of untrusted connection or cloud service provider. 

According to Mireshghallah \etal~\cite{mireshghallah_privacy_2020}, threats against the privacy are divided into two main groups: direct exposure threats and inferred exposure threats. In direct threat the attacker tries to gain access to the sensitive information directly while in the inferred ones, the attacker infers from the available information which does not contain the sensitive information. In most of the cases, they benefit from deep learning methods for the inference attacks~\cite{liew_faceleaks_2020,  salem_updates-leak_2019, song_analyzing_2020}.

\begin{figure}
\centering
\includegraphics[scale=0.265]{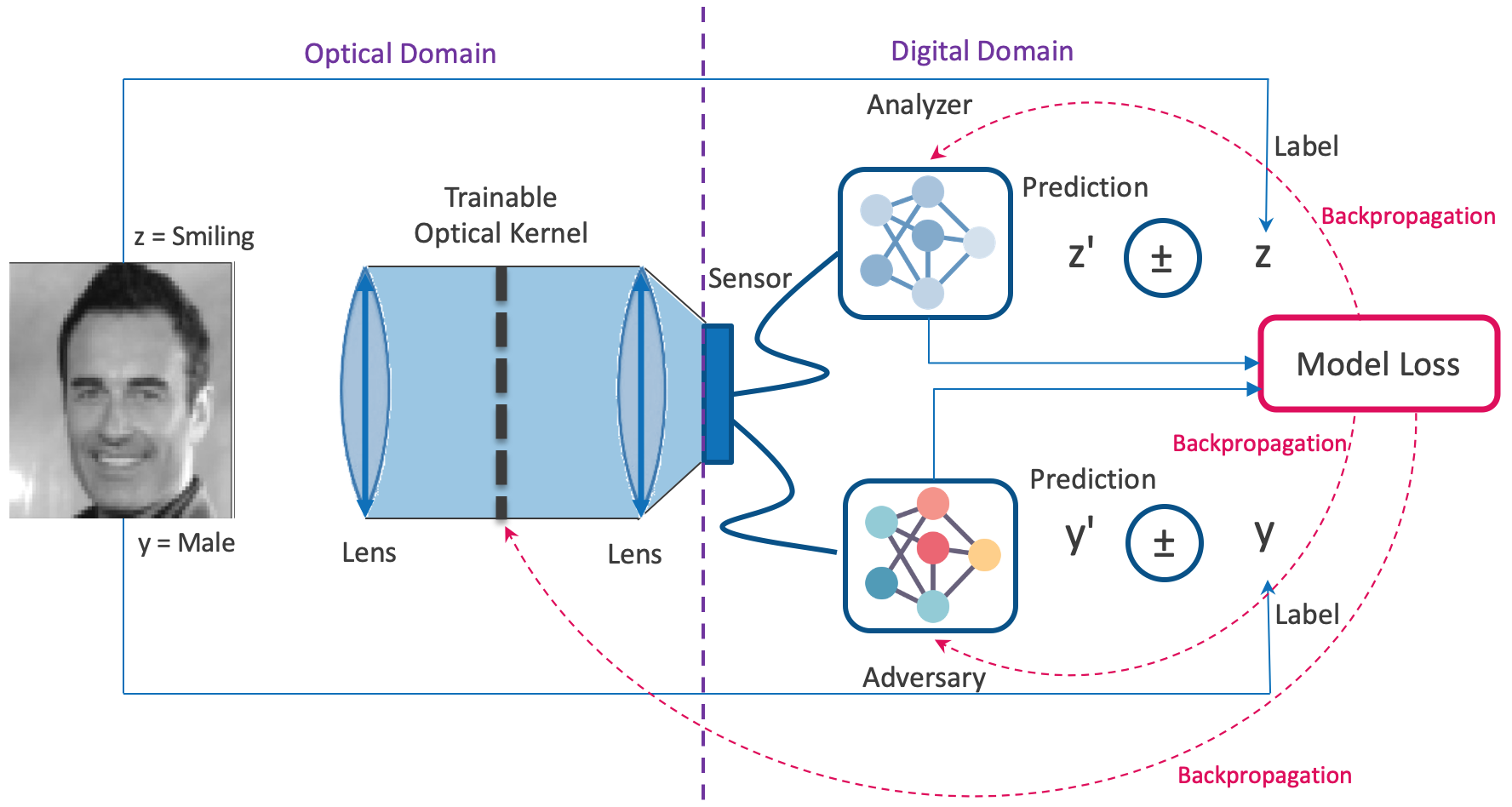}
\caption{A schematic view of the optical kernel and the training setup. The analyzer classifies the desired attribute while the adversary attacks to extract the sensitive content. The optical kernel, the digital analyzer, and the adversary are trained.}
\label{fig:setup}
\end{figure}

In this study, we propose a trainable optical front-end as the privatizer in order to perform the privacy-preserving task in the optical domain before the data reaches the image sensor and therefore the digital domain (see Figure \ref{fig:setup}). Thus, with this optical front-end, the system is invulnerable against any attack in the digital domain (edge or cloud). We pursue a data-driven approach to optimize the optical front-end to let pass the predefined desired information while removing the sensitive one. Also, since this processing happens passively in the optical domain, it has no memory and computational cost for the edge device which saves energy and reduces latency on such devices with confined processing power. As it will be shown, it is able to reduce significant amount of sensitive information from the face images without losing a considerable amount of desired content.

This approach is ideal for use-cases like camera-based feedback systems for smart TV content providers or billboard advertisements to recognize the satisfaction while not having access to the sensitive identity of users.  As this method is implemented in the optical domain without additional burden, it can be even further combined with methods of fully-digital privacy-preserving ecosystem to provide new features such as being resilient against direct attacks and increasing the total sensitive information blocking.

In the next sections, we discuss the related works (Section \ref{section:related}). Then we explain the proposed idea of a privacy-preserving trainable optical kernel (Section \ref{section:hybrid_optical}). In Section \ref{section:strategy}, we focus on two different methods to train this optical front-end. We demonstrate the performance of the proposed scheme in Section \ref{section:evaluation}. In the end, we summarize the paper and suggest some potential future works.
\vspace{5pt}
{\section{Related Works}\label{section:related}}
In this section, we categorize the related works into three parts and discuss their abilities and shortages.\vspace{5pt} 
{\subsection{
Fully-digital Privacy-Preserving Networks}\label{subsection:fully-dig}}

To alleviate the problem of users' privacy leakage, researchers implemented different methods of privacy-preserving neural networks. Mirjalili \etal~\cite{mirjalili_semi-adversarial_2018} proposed the idea of privacy-preserving semi-generative adversarial network which is an adversarial method of face recognition which neutralizes gender. Osia \etal~\cite{osia2018deep} investigated deep private feature extraction which divides the neural network between the edge device and cloud server. In this hierarchical method, they have a feature extractor part in the end device which keeps and sends the useful information while filtering out the sensitive attributes. They benefit from a specific loss function that has a cross-entropy term for the desired attribute inference, and a contrastive term for the sensitive content which results in a high accuracy of inference for desired content and a low amount of sensitive content leakage from the edge to the cloud. Additionally, Huang \etal~\cite{huang_generative_2019,huang_context-aware_2017} proposed the idea of an adversarially trained method with a privatizer neural network which sanitizes the input images from the sensitive content and an adversary which tries to infer the sensitive information. They evaluated their method on GENKI dataset~\cite{whitehill_discriminately_2011} which consists of face images and facial attributes and sanitized the gender information. 

These methods are only helpful to avoid the direct and inference attacks which are made on the information communicated to the cloud server. However, the attacks can also happen directly on the acquired images and the sensitive information which are stored on the edge devices. These methods cannot prevent these types of hacking access attacks since they all need to store the image data on the edge device to process and privatize it before sharing it with the cloud. Moreover, they are generally computationally intensive with numerous layers and do not consider the fact that the privatizer part should mainly be implemented on the user's edge devices with restricted power and storage~\cite{huang_generative_2019}. Although these methods do not contain any optical parts, their training strategy is inspirational for our purpose.
\vspace{5pt}
{\subsection{Hybrid Networks and Optical Convolution}\label{subsection:hybrid}}

The idea of dividing a deep neural network between the optical domain and the digital domain is generally a new method that shows promising results. Chang \etal~\cite{chang_hybrid_2018} proposed a two-layer convolutional neural network that has optimized diffractive optical filter as a convolution in the optical domain and a fully connected layer in the digital domain. The optical convolution showed an improvement in the classification accuracy without adding any computational burden. Lin \etal~\cite{lin_alloptical_2018} made an all-optical deep neural network with a set of 3D-printed diffractive layers. 

These studies were all done using the phase-varying optical masks which are usable for monochromatic and coherent light. Pad \etal~\cite{Pad_2020_CVPR} furthered this work by benefiting from an amplitude varying mask to implement a large optical convolution kernel which is able to work with polychromatic and incoherent light available in the natural scenes. They showed two orders of magnitude reduction in the computational cost while keeping the state-of-the-art accuracy.

The main purpose of these methods is to benefit from the optical-digital neural networks to reduce the memory footprint, the computational cost, and the power consumption of the neural network. In this study, we propose the idea of using an amplitude varying mask to remove the sensitive content while keeping the desired information which also has the above-mentioned advantages. \vspace{5pt}
{\subsection{Optical Privacy-Preserving Networks}\label{subsection:optical}}

The idea of benefiting from optical devices to provide privacy-preserving vision systems has been implemented by different methods. Nakashima \etal~\cite{nakashima_development_2010} used a cylindrical lens to preserve privacy in people's position and movement detection. Pittaluga and Koppal~\cite{pittaluga_pre-capture_2017} exploited a defocusing blur method and an optical k-anonymity scheme based on averaging of the target face image on $k-1$ neighbor images in the dataset. However, their method of anonymization needs precise calibration and to have access to more than one image from the dataset during the inference phase which is not well-aligned with the idea of privacy-preserving vision systems. Wang \etal~\cite{wang_privacy-preserving_2019} proposed the idea of preserving privacy with a coded aperture for action recognition. In their experiments, they used pseudorandom optical mask which almost has uniform distribution and they exploited phase correlation to perform action recognition, without being aware of the mask design. 

In contrast, in this study, we propose a method which uses a trainable mask that can be counted as the first layer of neural network. This is the first hybrid optical digital neural network for privacy-preserving applications which is fully trainable and data-driven. To the best of our knowledge, no work has investigated using a trainable part in the optical domain to improve the privacy-preserving property. \vspace{5pt}
{\section{Training an Optical Kernel for Preserving Privacy}\label{section:hybrid_optical}}

\begin{figure}
\centering
\includegraphics[scale=0.18]{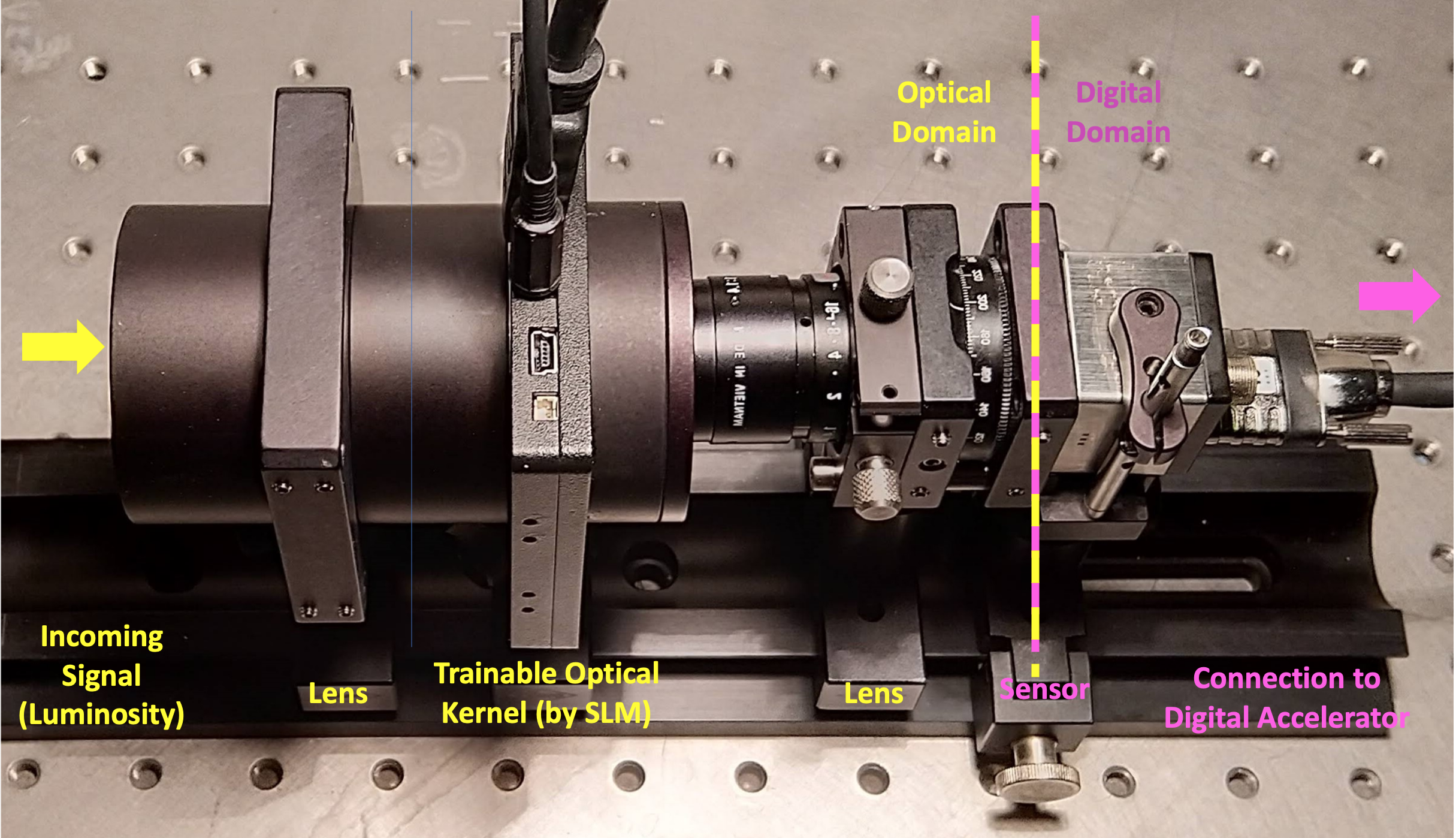}
\caption{Prototype optical setup. The trainable optical kernel is implemented by an SLM.}
\label{fig:real_setup}
\end{figure}

In most computer vision tasks such as facial attribute detection, action recognition, or body-pose estimation, a standard image of the person is taken by a camera with a conventional optical front-end (a lens system) and fed into a digital neural network. In this scheme, a quite high amount of private data like the identity of the person or the gender is automatically shared with the digital system. This data is intrinsically irrelevant to the task and entering them into the digital domain is the Achilles' heel for privacy attacks. In fact, once this data reaches the digital domain, it provides the possibility for different types of attackers to access it. Thus, the best way to preserve privacy in such tasks is to remove the sensitive data in the optical domain before reaching the image sensor. In doing this, we benefit from a relatively simple optical front-end proposed in ~\cite{Pad_2020_CVPR} which performs the convolution of the scene with a pretrained amplitude varying mask in the optical domain. Figure \ref{fig:setup} shows a schematic view of the privacy-preserving approach and  Figure \ref{fig:real_setup} shows its realization.

The relation between the signal reaching the image sensor $I(\mathbf{x})$, the scene $J(\mathbf{x})$, and the optical kernel $K(\mathbf{x})$ is
\begin{equation}\label{eq:mainconv}
I(\mathbf{x}) = \left(J(\alpha~\cdot)*\gamma K(\gamma~\cdot)T(\gamma~\cdot)\right)(-\mathbf{x})
\end{equation}
where $\mathbf{x}\in\mathbb{R}^2$ is the pixel position variable, $*$ denotes the standard $2$-dimensional convolution, $T(\mathbf{u})$ is a radial function that is in practice very close to constant $1$. Also, scalars $\alpha$ and $\gamma$ depend on the physical parameters of the system such as the lens powers and the distances between the optical components. In practice, the kernel $K(\mathbf{x})$ can be implemented by a Spatial Light Modulator (SLM)~\cite{ambs_spatial_2007} which is a programmable transmissive mask (Figure \ref{fig:real_setup}) or micro-fabrication~\cite{orabona_photomasks_2013} and aerosoljet  printing~\cite{wilkinson2019review} which result in a fixed mask but with much higher resolution and miniature size. Also, when high precision is not required, inkjet printing on transparent substrates is a low-cost option \cite{singh2010inkjet}.

As we see, in this setup, the result of the convolution that reaches the image sensor is not the raw image. Our general idea is that by appropriately training the optical kernel, we will be able to sanitize the data from sensitive contents while letting the useful information pass through. Note that as this processing happens in the optical domain, regardless of the number of parameters of the optical kernel, it imposes no additional computational cost to the acquisition system.

We perform two different strategies for training this optical kernel, the first one is based on Generative Adversarial Privacy (GAP)~\cite{huang_context-aware_2017,huang_generative_2019}, and the other one is based on an Inverse Siamese architecture (IS)~\cite{osia2018deep}. We consider the so-called post-training adversary which is an attacker network that freely trains on the feature set generated by the optical front-end to extract the sensitive content. We use this as a powerful tool to evaluate the effectiveness of the proposed privacy-preserving method.

We demonstrate the performance of the proposed method by experiments on different attributes of the CelebA dataset~\cite{liu2015faceattributes}. In addition, we demonstrate the performance of the deep visualization attacker which tries to reconstruct the original face based on the optical features.\vspace{5pt}
{\section{Training Strategies}\label{section:strategy}}
In this section, we focus on two training strategies that are commonly used for privacy-preserving tasks, and explain how to utilize them to train the proposed hybrid optical-digital neural network.\vspace{5pt}
{\subsection{Training by Generative Adversarial Privacy (GAP)}\label{subsection: GAP}}

\begin{figure}
\centering
\includegraphics[scale=0.205]{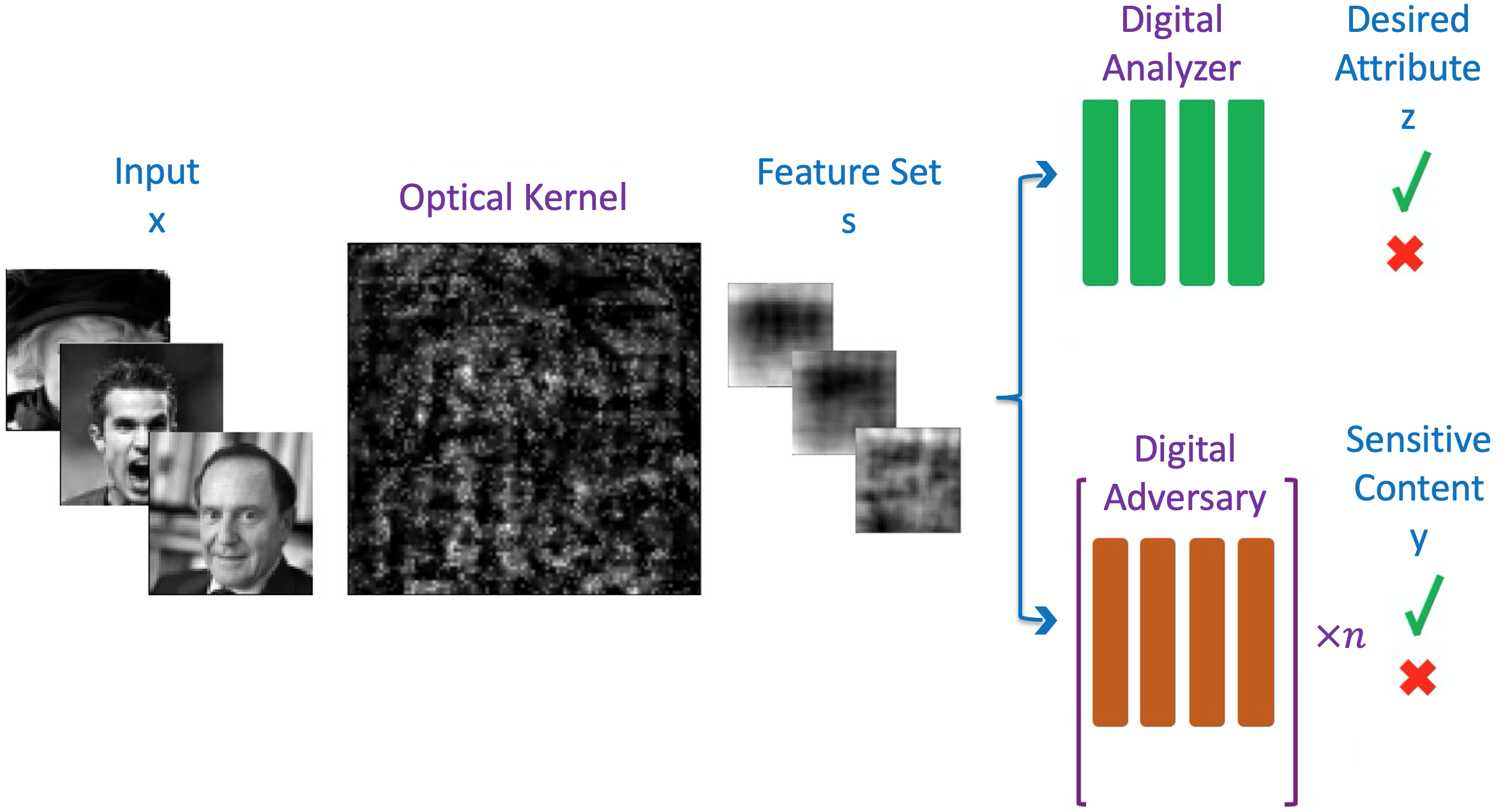}
\caption{GAP strategy for training the optical kernel along with the analyzer and the adversary networks. The optical kernel receives the input $x$, extracts the feature set $s$, and feeds it into the digital analyzer and adversary. For each update of the analyzer, the adversary is updated $n$ times.}
\label{fig:tr_step_1}
\end{figure}

Inspired by \cite{huang_context-aware_2017,huang_generative_2019}, we propose a generative adversarial privacy (GAP) scheme to train the optical kernel. The optical kernel works as a feature extractor which generates feature sets and sends them to the digital domain. In the digital domain, there is a network called \textit{analyzer} which tries to provide a high accuracy for the desired content classification. On the other hand, there is an \textit{adversary} network that tries to provide a high accuracy for the sensitive content extraction, see Figures \ref{fig:setup} and \ref{fig:tr_step_1}.

In the forward pass, the input images $x$ is given to the neural network. The image is convolved with the optical kernel and the generated feature map $s$ is fed into the digital analyzer and digital adversary where the first one estimates the desired attribute $z$ and the second one estimates the sensitive content $y$.

In the backward pass, first, we update the parameters of the optical kernel $K(\cdot)$ which are denoted by $\theta$ and the digital analyzer's parameters $\phi$ based on the following optimization problem:
\begin{equation}
\begin{aligned}
    \theta^*,\phi^* =& \argminA_{\theta,\phi} \sum_{i} -z_i  \log(\hat{z}_i|x_i;\theta,\phi) \\ &
    -\lambda \Bigg(\sum_{i} -y_i   \log(\hat{y}_i|x_i;\theta,\psi)\Bigg)
\end{aligned}
\label{eq:1}
\end{equation}
where $z_i$ and $y_i$ are the labels of the desired attribute and the sensitive content corresponding to the input $x_i$, respectively. Also, $\hat{z}_i$ and $\hat{y}_i$ are the output predictions of the analyzer and the adversary, and $\psi$ is the parameters of the adversarial network. The scalar $\lambda$ is a hyper-parameter which aligns the power of the first term (desired attribute loss) and the second term (sensitive content loss). In this step, the parameters of the optical kernel ($\theta$) are updated such that they increase the accuracy of the analyzer network but decrease the accuracy of the adversarial network. Note that $\psi$ is fixed in this step.

After that, it is time for the adversary to be trained. The update of the adversary parameters happen based on the following optimization problem:
\begin{equation}
\begin{aligned}
    \psi^* =& \argminA_{\psi} 
     \sum_{i} -y_i \log(\hat{y}_i|x_i;\theta,\psi).
\end{aligned}
\label{eq:2}
\end{equation}
In this step, the optical kernel as well as the analyzer are fixed and only the adversary is updated. Indeed, for each update of the optical kernel and analyzer parameters, we perform $n$ updates of the adversary parameters. The value of $n$ is another hyperparameter that needs to be set.

\begin{figure}
\centering
\includegraphics[scale=0.205]{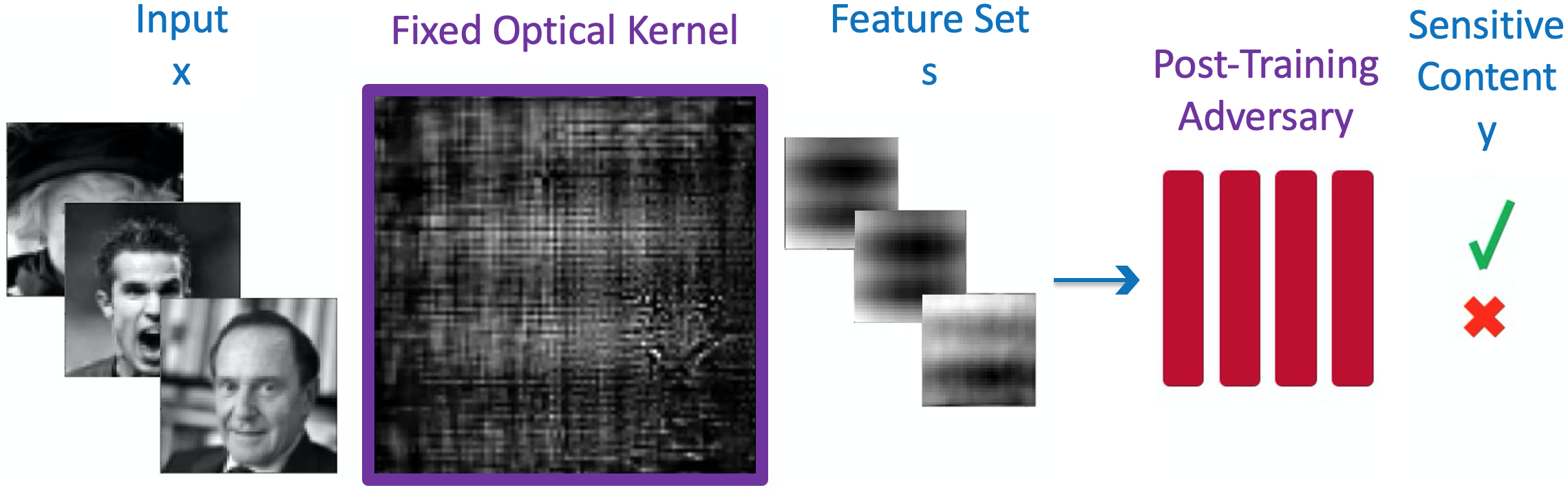}
\caption{Training of the attacker neural network which is also called as post-training adversary. After training of the optical kernel, it is kept fixed and then the attacker is trained to extract the sensitive content from the output features of the optical domain.}
\label{fig:tr_step_2}
\end{figure}

After finishing the training phase, the optical kernel is fixed and we let an attacker neural network train itself on the extracted features from the dataset to infer the sensitive content. We call this attacker neural network \textit{post-training adversary} and assume that it has white box access (even knowing the optical kernel exactly) to the trained optical kernel, see Figure \ref{fig:tr_step_2}. The optimization problem to find the post-training adversary parameters $\zeta$ is
\begin{equation}
\begin{aligned}
    \zeta^* =& \argminA_{\zeta} 
     \sum_{i} -y_i \log(\hat{y}_i|x_i;\theta,\zeta)\\=& \argminA_{\zeta} 
     \sum_{i} -y_i \log(\hat{y}_i|s_i;\zeta).
\end{aligned}
\label{eq:3}
\end{equation}
It is worth mentioning that in order to keep the comparisons fair, the number of epochs are the same for the analyzer and the post-training adversary.

\vspace{5pt}
{\subsection{Training by Inverse Siamese (IS)}\label{subsection: Siamese}}

\begin{figure}
\centering
\includegraphics[scale=0.20]{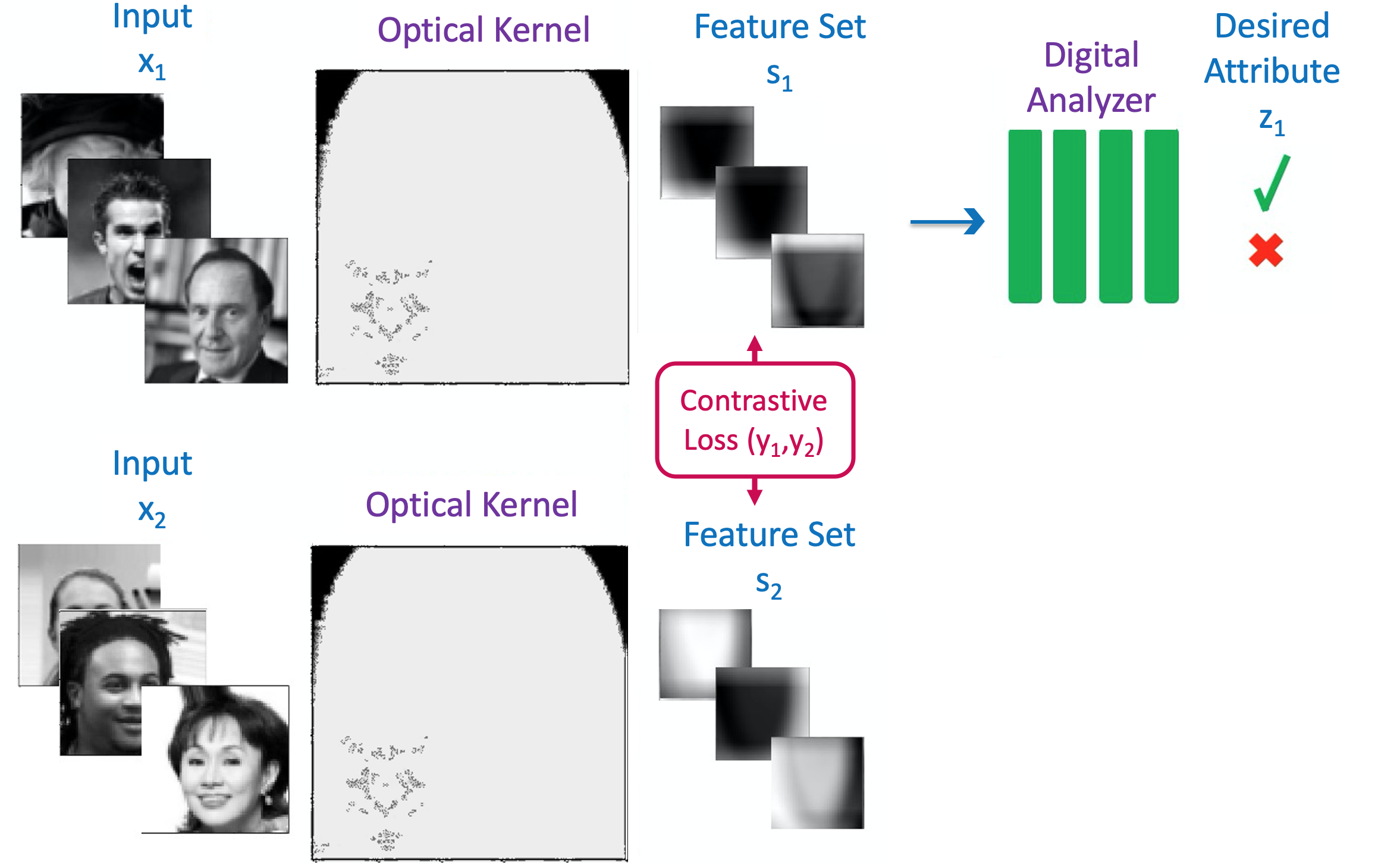}
\caption{IS strategy for training the optical kernel and the analyzer. The optical kernel receives two input images from the training mini-batch. To sanitize the sensitive content, the generated feature sets are compared with each other and if the sensitive label is the same, they are distanced by the contrastive loss. Otherwise, they are pushed toward each other.}
\label{fig:tr_inv_siamese}
\end{figure}

The second method to train the optical kernel is called the Inverse Siamese method (IS). This method is firstly proposed by~\cite{osia2018deep} to train the deep feature extractors and here we implement and modify it to train an optical convolution. Figure \ref{fig:tr_inv_siamese} shows an overview of this method.

In this method, there is again an analyzer that tries to provide a good accuracy for the inference of the desired attribute of the input images by jointly training itself and the optical convolution using the cross-entropy loss. However, to sanitize the generated feature map $s$ from the sensitive content, at each training iteration, two inputs $x_i$ and $x_j$ are selected from the training mini-batch and will be convolved with the optical kernel. After performing the convolution, the two generated feature sets $s_i$ and $s_j$ are compared to each other based on their sensitive attribute labels $y_i$ and $y_j$. It is done by a contrastive loss term that is added to the cross-entropy loss of the desired attribute prediction (Eq. \ref{eq:4}). The loss is designed such that it pushes the feature maps away from each other when the sensitive attributes are equal and pushes them toward each other when the sensitive attributes are different. In fact, this is the reason why the method is called inverse Siamese as in the Siamese networks the feature maps of the data points with the same labels are pushed toward each other and vice versa~\cite{osia2018deep}. Therefore, the optimization problem for training the network with IS strategy is
\begin{align}\label{eq:4}
     \theta^*,\phi^* =& \argminA_{\theta,\phi}  \sum_{i} -z_i \log(\hat{z}_i|x_i;\theta,\phi) \\ &
    -\lambda \sum_{B\in\mathcal{B}}\frac{1}{|B|} \Bigg(\sum_{\begin{smallmatrix}i,j\in B\\y_i=y_j\end{smallmatrix}} \left\|(s_i|x_i;\theta) -(s_j|x_j;\theta)\right\|_2 \nonumber\\ &~~~~~~~~~~~~~~~~~~~~~~~~~ -\sum_{\begin{smallmatrix}i,j\in B\\y_i \neq y_j\end{smallmatrix}}\left\|(s_i|x_i;\theta) -(s_j|x_j;\theta)\right\|_2\Bigg)\nonumber
\end{align}
where $\mathcal{B}$ is the set of all batches and $|B|$ indicates the number of elements of batch $B$. 
Again, after training the optical kernel and the analyzer network, the post-training adversary is trained based on Eq. \ref{eq:3}.\vspace{5pt}
{\section{Evaluation and Experiments}\label{section:evaluation}}

To evaluate the performance of the proposed privacy-preserving scheme, we performed experiments on CelebA dataset~\cite{liu2015faceattributes} which contains $202599$ face images of celebrities divided into $162770$ training samples ($\approx 80 \%$), $19867$ validation samples $(\approx 10 \%)$, and $19962$ testing samples ($\approx 10 \%$). Each of the images has 40 different binary attributes in this dataset. The original images have a size of $178 \times 218$ and for our experiments, we center-cropped these images into $170 \times 170$ and downsampled them into $64 \times 64$. Also, the images are grayscaled. The reason is the fact that the proposed optical convolutions cannot have more than one channels in the optical domain. However, it is possible to manufacture kernels for different wavelengths of light and also using color image sensors which are beyond the scope of this study and left for future investigations. 

In our illustrative experiments, we choose the \textit{smiling} and \textit{mouth slightly open} attributes as the desired attributes and the \textit{gender} and \textit{wearing makeup (lipstick)} attributes as the sensitive content. It is important to notice that the dataset is balanced for the smiling attribute, but $61.3\%$ of the dataset is female. For \textit{wearing lipstick} attribute the data is divided as $47.2\%-52.8\%$ and for \textit{mouth slightly open} as $48.3\%-51.7\%$. It is worth mentioning that hiding the gender attribute is harder than hiding the identity and if the model is able to do this, it automatically ends in removing the identity information to a large extent.\vspace{5pt}

{\subsection{Training and Validation}\label{section:training}}

The resolution of the optical convolution is set to $100 \times 100$ where its weights are clamped between $0$ and $1$. On the extremes, the value $0$ means that no light is passing through that point on the optical kernel and the value $1$ means that all of the incoming light passes through that point. Unlike the conventional convolutional layers, here the convolution kernel is bigger than the input size, thus we need to pad the input quite largely to cover all the kernel. It is padded by $50$ on each side. Then, the pixel pitch of the image sensor is designed such that it is equivalent to the stride equal to $2$. Consequently, the output size of the optical domain which is also the input of the digital domain has a size of $32 \times 32$. Notice that here is a trade-off since a higher resolution of the feature map on the image sensor can result in a higher accuracy for detecting the desired attribute but concomitantly it might result in higher information leakage for the attacker to extract the sensitive content.
\begin{figure*}[!t]
\begin{center}
\includegraphics[scale=0.46]{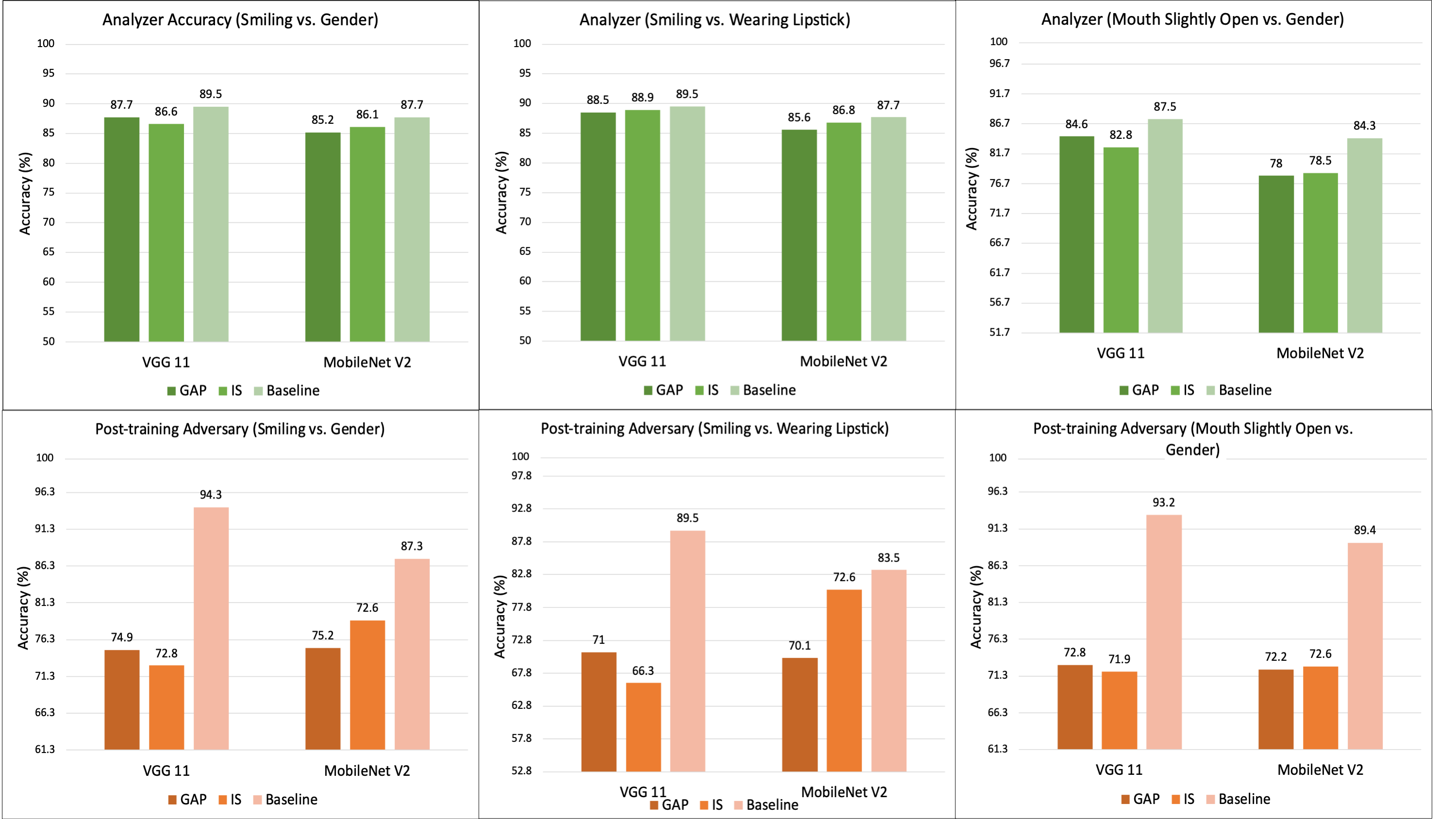}
\end{center}
   \caption{The accuracy of desired content analyzer vs. sensitive content attacker for the different experiments on different attributes. In each case, the accuracy of the privacy-preserving methods trained by GAP and IS strategies are compared with the corresponding baseline. The goal is to have a low reduction in accuracy of the analyzer and a high reduction in accuracy of the attacker. The lowest value displayed on the vertical axis corresponds to the lowest possible accuracy of the trivial classifier (see Section \ref{section:testing})}.
\label{fig:charts}
\end{figure*}

To demonstrate the adaptability of our trainable optical privacy-preserving method to different network architectures, two standard neural networks of VGG11~\cite{simonyan_very_2015} and MobileNet V2~\cite{sandler_mobilenetv2_2019} have been implemented for the analyzer and adversaries. The selected networks are not the most complex ones and were chosen with the envisaged application on edge devices with confined memory and computational power. The original architectures are slightly modified due to the specific number of input and output channels. The number of input channels is $1$ as the input image is gray-scale (see the beginning of this section) and the optical kernel is equivalent to a single filter, but with a huge number of parameters.  There are $2$ output channels to perform the binary classification of the attributes. As mentioned, the selected experiments are \textit{smiling} vs. \textit{gender}, \textit{smiling} vs. \textit{wearing lipstick}, \textit{mouth slightly open} vs. \textit{gender}. 

The GAP strategy has been trained for $100$ epochs. The coefficient of the privacy-preserving terms of loss ($\lambda_{\text{GAP}}$) and the number of steps for adversary training ($n$) have been tuned based on the validation set for each experiment (Eq. \ref{eq:2}). We used the Adam optimizer with a learning rate of $0.002$ for all the analyzers, adversaries, and post-training adversaries. After finishing the training, the optical kernel is fixed and the post-training attacker is trained to achieve the highest possible accuracy for sensitive content inference (Eq. \ref{eq:3}). In the IS strategy, the networks were also trained for $100$ epochs. The coefficient of the privacy-preserving term of loss ($\lambda_{\text{IS}}$) was tuned based on the validation set for each case. Size of the Siamese mini-batch was set to $|B|=32$ and it is the same for all batches (Eq. \ref{eq:4}). Again the Adam optimizer with the learning rate of $0.002$ was used for both the analyzer and the post-training adversary. After finishing the training, the post-training attacker has been trained on the extracted features by the trained optical kernel (Eq. \ref{eq:3}).

\begin{figure*}
\begin{center}
\includegraphics[scale=0.25]{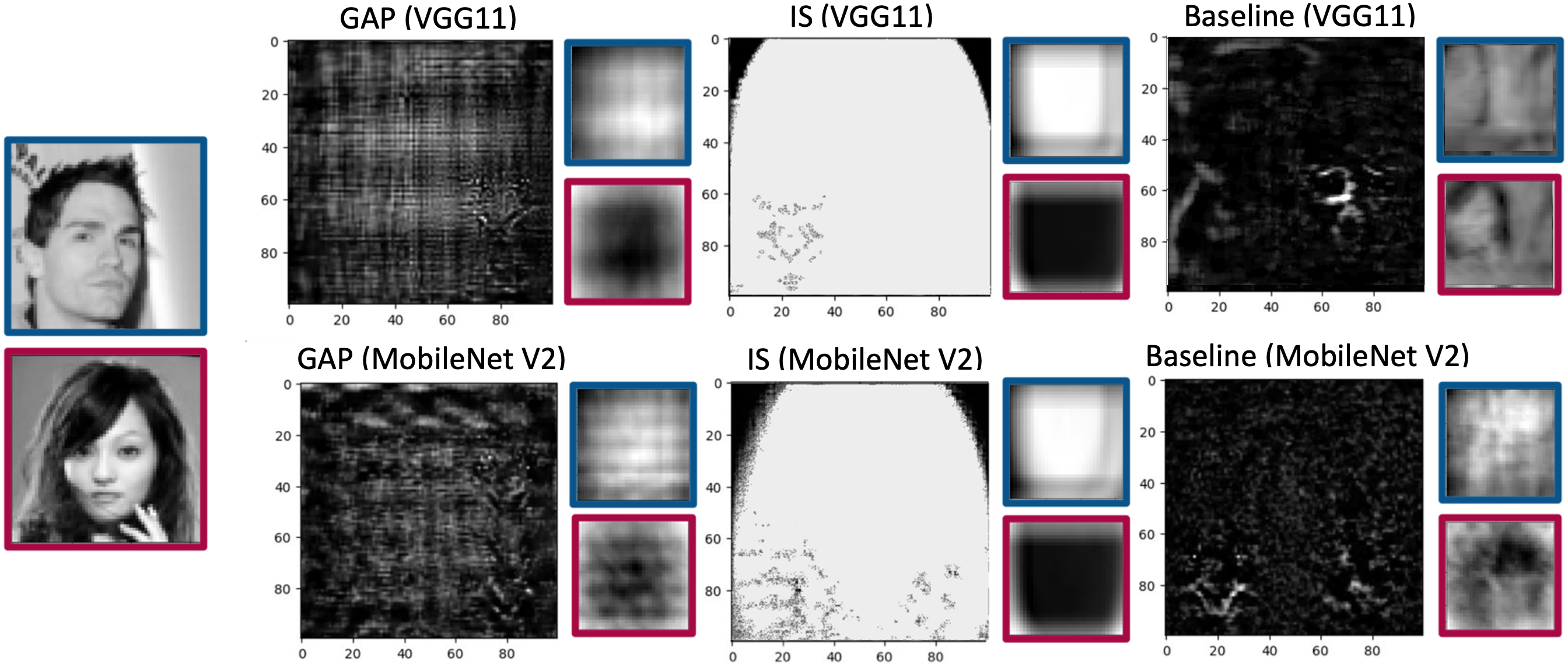}
\end{center}
   \caption{The optical kernel trained for different architectures and training strategies for the smile vs. gender experiment. On the left side, the two input images are shown. On the right side, the kernels and the corresponding feature sets are demonstrated.}
\label{fig:input-feature}
\end{figure*}
\vspace{1pt}
{\subsection{Testing Performances and Visualization}\label{section:testing}}
After finishing the training phase, the networks have been tested on the testing subset of CelebA dataset. The results are reported in Figure \ref{fig:charts}.

In order to have a clear understanding of the effectiveness of the proposed privacy-preserving method in the experiments on the different attributes and the training strategies, we need to have the lower and upper bounds of the achievable detection accuracies of the desired attribute and the sensitive content. The lower bound is the percentage of the label that has the majority in the dataset. It is in fact the accuracy of the trivial detector that blindly outputs the label of the majority class. In our case, the dataset is balanced for smiling attribute and thus the lower bound of its accuracy is $50\%$. However, $61.3\%$ of the images are female, $51.7\%$ have closed mouth, and $52.8\%$ do not wear lipstick. Thus, the lower bounds for them are $61.3\%$, $51.7\%$, and $52.8\%$ respectively. The importance of these values is twofolds: firstly, it means that for instance, we cannot expect that any privacy-preserving method results in an accuracy less than $61.3\%$ on the gender; secondly, we understand that how close the accuracy of the proposed privacy-preserving attribute detector is to its trivial counterpart. The lowest level of the charts in Figure~\ref{fig:charts} indicates these values.

To obtain the upper bound of the accuracy, baseline networks have been made and trained for VGG11~\cite{simonyan_very_2015} and MobileNet V2~\cite{sandler_mobilenetv2_2019} architectures. These baselines have the same optical-digital architectures as the privacy-preserving ones; however, during training, no privacy-preserving strategy like GAP or IS is exploited. The analyzer baseline is just an optical-digital model that is trained for the desired information classification. After finishing the training, an attacker is made by training the digital domain part of the neural network while keeping the optical kernel parameters frozen. Indeed, the baseline shows two facts: firstly, the maximum accuracy that we can expect from the privacy-preserving analyzer network; secondly, it shows the amount of sensitive information leakage we will have from the optical domain, when we train the hybrid model only considering the classification accuracy of the desired useful content. As an example, as we see in Figure \ref{fig:charts}\textcolor{red}{-Top-Left}, the accuracy of smile detection for the baseline hybrid VGG11 and MobileNet V2 networks are $89.5\%$ and $87.7\%$, respectively. Thus, after incorporating the privacy-preserving strategies in the training phase, we cannot expect the analyzer networks to achieve accuracies higher than these numbers. We also see in Figure \ref{fig:charts}\textcolor{red}{-Bottom-Left} that without privacy-preserving strategy, the gender detector attacker achieves the accuracy of $94.3\%$ and $87.3\%$ when using the optical kernels trained for the baseline smile detection with VGG11 and MobileNet V2 networks, respectively.

\begin{figure}
\begin{center}
\includegraphics[scale=0.2]{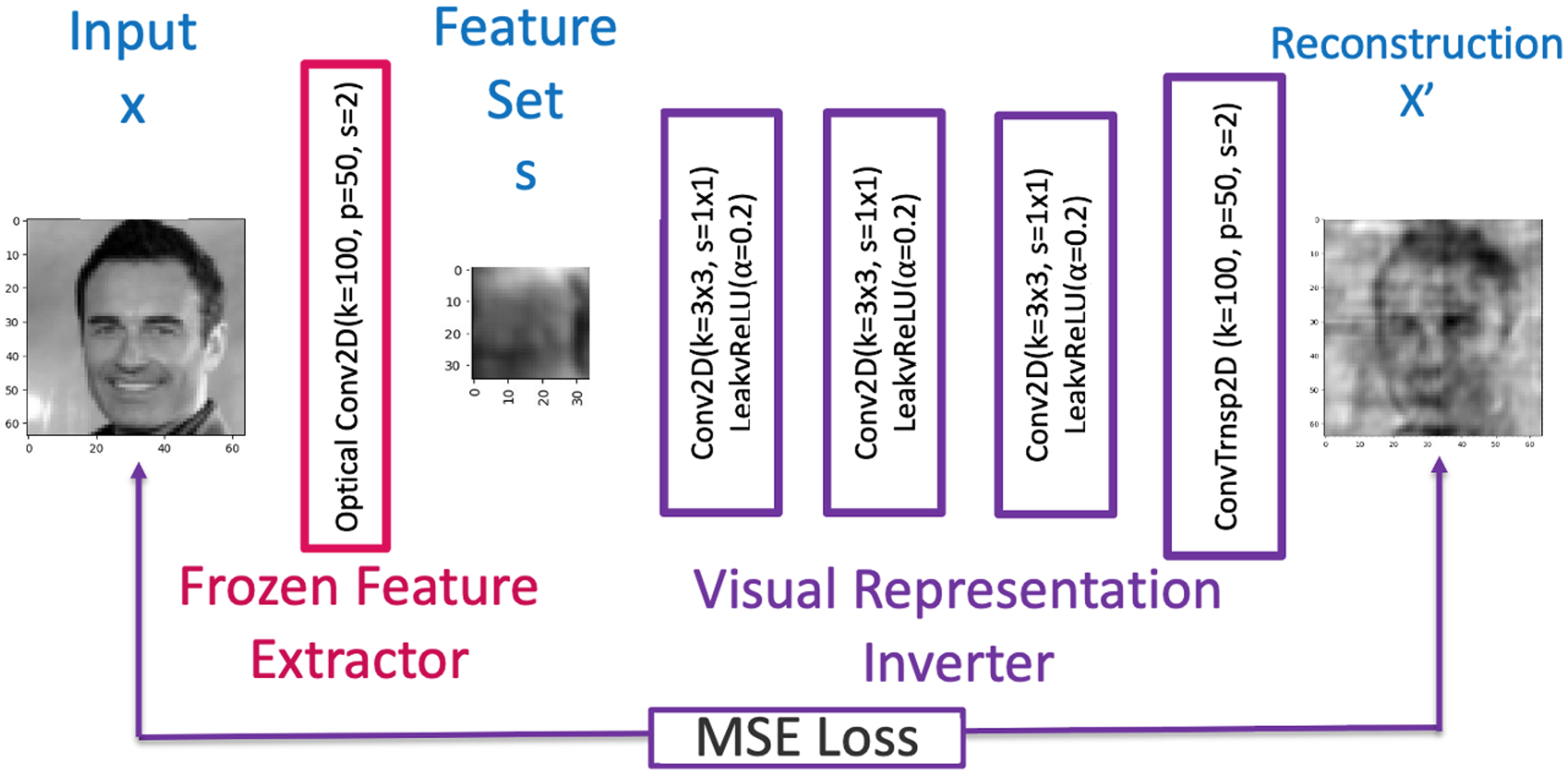}
\end{center}
   \caption{Left: The visual representation inverter, based on the idea of~\cite{krizhevsky_imagenet_2017}. It uses the pixel-wise MSE loss to train the decoder part. After finishing the training the trained reconstructor is used to reconstruct the images from the extracted feature sets.}
\label{fig:recon}
\end{figure}

Now, Figure \ref{fig:charts}\textcolor{red}{-Bottom-Left} shows that by training the VGG11 hybrid network with the GAP privacy-preserving strategy, the accuracy of the gender detector attacker is reduced by $19.4\%$ which is equal to a $58.8\%$ of the sensitive content. This happens while the smile detector loses only $1.8\%$ of its accuracy which is equal to $4.5\%$ of the desired content according to Figure \ref{fig:charts}\textcolor{red}{-Top-Left}. Also, by training the same network with the IS privacy-preserving strategy, the accuracy of the attacker is decreased by $21.5\%$ equal to $65.1\%$ of the sensitive information while we lose only $2.9\%$ of the accuracy of the analyzer ($=7.3\%$ of desired content). The results for the MobileNet V2 on smile detection vs. gender and also the other experiments on the other attributes are available in Figure \ref{fig:charts}.

As shown in Figure \ref{fig:charts}, both training strategies generally provide a significant reduction in the sensitive information without considerably affecting the desired information. Comparing them together, IS approach is able to achieve higher sensitive content removal in some cases (like the 3 experiments based on VGG11); however, GAP is more robust against the choice of the neural network, and desired and sensitive attributes.

  
   
    
    


\begin{figure*}
\begin{center}
\includegraphics[scale=0.28]{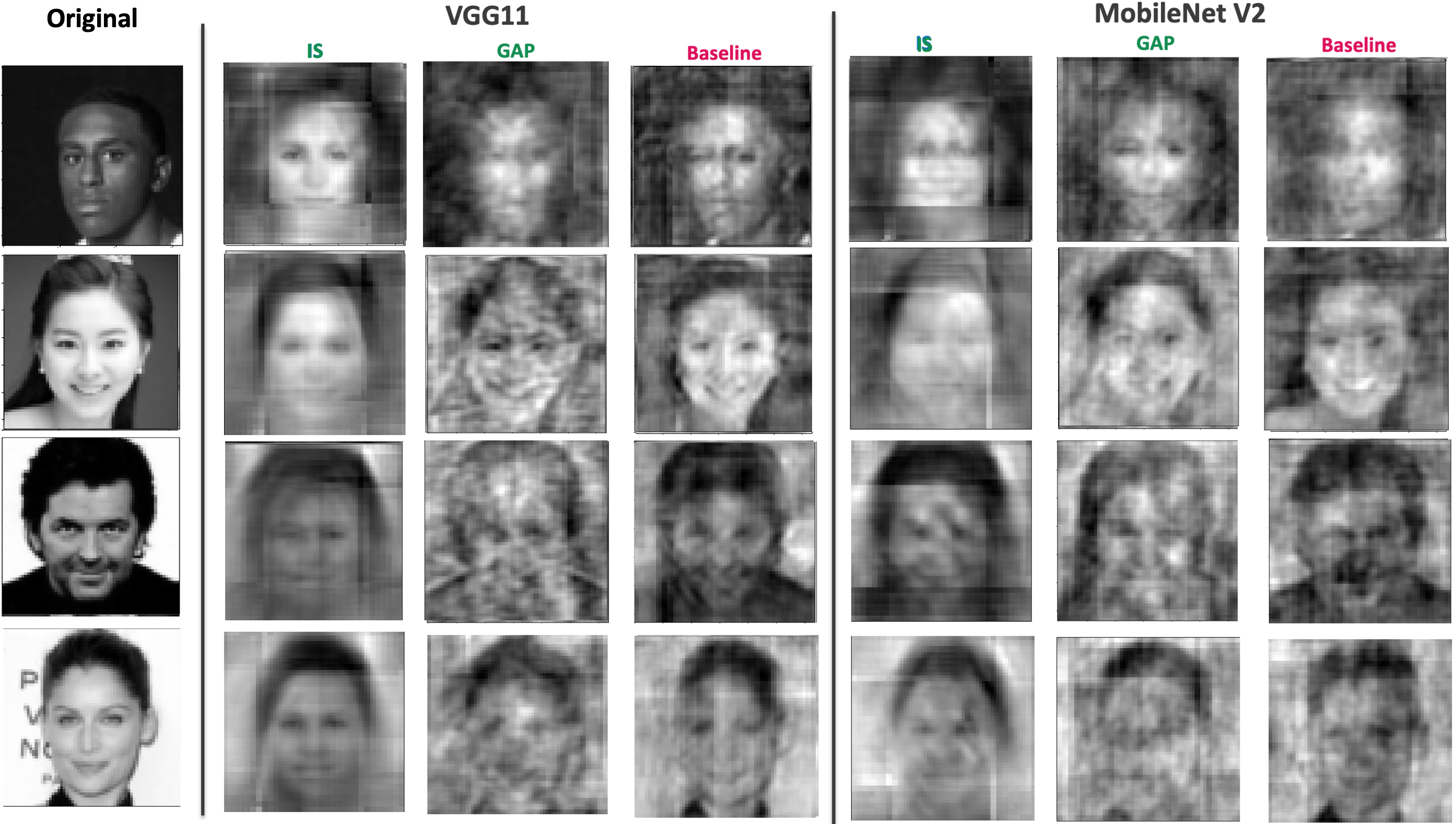}
\end{center}
   \caption{Examples of input and reconstructed images generated by the visual representation inverter from the extracted features of optical kernels made by different architectures and training strategies for the smile vs. gender experiment.}
\label{fig:input-recon}
\end{figure*}

To have an intuitive understanding of the privacy-preserving performance of the optical kernel, we conduct another experiment. First, we look at the examples of signals that we receive on the image sensor. Figure \ref{fig:input-feature} shows two sample images from the test set, the optical kernels which are shown were generated by different training strategies for the smile vs. gender experiment. The output of optical domain which is indeed the input of the digital domain is also shown. As we see, there is no human identifiable information in the 2-dimensional signal of the image sensor.

In the second step, we train a reconstructor neural network that aims at reconstructing the original face based on the signal on the image sensor. This can be seen as the reconstruction attack to extract the sensitive content. To do so, we used a method presented by Dosovitskiy and Brox~\cite{dosovitskiy_inverting_2016}. They proposed their method for inverting feature representations of networks like AlexNet~\cite{krizhevsky_imagenet_2017} and it is successful for different layers. We used their idea to reconstruct the original face from the extracted optical features. The network we made based on their approach is shown in Figure \ref{fig:recon}.


In this method, the trained optical kernel is used as a fixed encoder of an autoencoder where the visual representation decoder is its decoder. In the decoder, there are three convolutions which do not change the size of input. After that there is an upconvolution (transposed convolution) which has the same features as the optical convolution and generates the reconstructed image. This inverter is trained using a pixel-wise mean squared error between the original and the reconstructed image.  After training, the visual representation inverter can be used to reconstruct the original images from the output of the optical convolution. Figure~\ref{fig:input-recon} contains examples of input and reconstructed images for different architectures, training strategies, and baselines for the experiment of smile vs. gender.

As we see, recognizing the sensitive attributes like the gender is not possible in the majority of cases from the reconstructed images when the privacy-preserving optical kernel is used. Detecting the identity is even harder and it is impossible for most of the cases.  However, these tasks are generally doable from the baseline reconstructions.\vspace{5pt}
{\section{Conclusion and Future Works}\label{section:conclusion}}

In this study, we presented a privacy-preserving vision system based on a trainable optical kernel. This system is robust against the direct access and hacking attacks since the sensitive content is removed in the optical domain and is not loaded in the digital domain at all. Moreover, as this processing happens in the optical domain, it does not add any computational or memory burden to the system. To the best of our knowledge, it is the first optical privacy-preserving method that is trainable and works as a part of the neural network. We studied two different strategies of training for this optical privatizer and compared them with the baselines. Several experiments showed that the trainable optical kernel keeps the detection accuracy of the different desired attributes high while substantially reducing the amount of sensitive content leakage. For example, it is able to reduce the gender content by $65.1\%$ while just losing $7.3\%$ of the smiling feature as the desired attribute. We showed that the method is generic and adaptable to different digital neural networks.

Future work includes extending this work to color images. Also, from a theoretical point of view, it is valuable to find the suitable regularization terms for the loss function or projection operators for update steps of training that constrain the search space of the optical kernel to the ones that are mathematically irreversible. Additionally, as this privacy-preserving method works in the new domain of optics, the combination of this idea with different available methods of digital privacy-preserving ecosystem, such as homomorphic encryption \cite{tsipouras_applying_2020} can be investigated.

{\small
\bibliographystyle{ieee_fullname}
\bibliography{main}
}

\end{document}